\documentclass{article}

\usepackage{arxiv}

\usepackage[utf8]{inputenc} 
\usepackage[T1]{fontenc}    
\usepackage{hyperref}       
\usepackage{url}            
\usepackage{booktabs}       
\usepackage{amsfonts}       
\usepackage{nicefrac}       
\usepackage{microtype}      
\usepackage{cleveref}       
\usepackage{lipsum}         
\usepackage{graphicx}
\usepackage{natbib}
\usepackage{enumitem}
\usepackage{doi}

\title{Question Answering for Decisionmaking in Green Building Design: A Multimodal Data Reasoning Method Driven by Large Language Models}

\author{ 
\normalfont
  \href{https://orcid.org/0000-0003-2018-2131}{\includegraphics[scale=0.06]{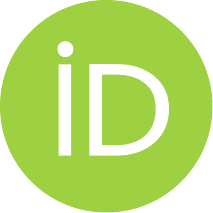}\hspace{1mm}\textbf{Yihui Li}*} \\
	School of Architecture\\
	Tsinghua University\\
	\href{liyihui23@mails.tsinghua.edu.cn}{\texttt{liyihui23@mails.tsinghua.edu.cn}} \\
	\And
    \normalfont
    \href{https://orcid.org/0009-0000-8474-3490}{\includegraphics[scale=0.06]{orcid.pdf}\hspace{1mm}\textbf{Xiaoyue Yan}}\thanks{These authors contributed equally to this work.} \\
        School of Architecture\\
	Tsinghua University\\
	\href{yanxy23@mails.tsinghua.edu.cn}{\texttt{yanxy23@mails.tsinghua.edu.cn}} \\
        \AND
        \normalfont
	\textbf{Hao Zhou} \\
        Center of Tsinghua Think Tanks\\
	Tsinghua University\\
	\href{haozhou2020@tsinghua.edu.cn}{\texttt{haozhou2020@tsinghua.edu.cn}} \\
	\and
    \normalfont
	\textbf{Borong Lin} \\
        School of Architecture\\
	Tsinghua University\\
	\href{linbr@tsinghua.edu.cn}{\texttt{linbr@tsinghua.edu.cn}} \\
}

\renewcommand{\shorttitle}{\textit{QA for DGBD: A Multimodal Data Reasoning Method Driven by Large Language Models}}

\hypersetup{
pdftitle={QA for DGBD: A Multimodal Data Reasoning Method Driven by Large Language Models},
pdfsubject={q-bio.NC, q-bio.QM},
pdfauthor={Yihui Li, Xiaoyue Yan, Hao Zhou, Borong Lin},
pdfkeywords={Design Thinking and Applied Research, Artificial Intelligence in Design, Green Building, Question Answering, Human-AI Interaction}}

\begin{document}
\maketitle

\begin{abstract}
In recent years, the critical role of green buildings in addressing energy consumption and environmental issues has become widely acknowledged. Research indicates that over 40\% of potential energy savings can be achieved during the early design stage. Therefore, decision-making in green building design (DGBD), which is based on modeling and performance simulation, is crucial for reducing building energy costs.

However, the field of green building encompasses a broad range of specialized knowledge, which involves significant learning costs and results in low decision-making efficiency. Many studies have already applied artificial intelligence (AI) methods to this field.

Based on previous research, this study innovatively integrates large language models with DGBD, creating GreenQA, a question answering framework for multimodal data reasoning. Utilizing Retrieval Augmented Generation, Chain of Thought, and Function Call methods, GreenQA enables multimodal question answering, including weather data analysis and visualization, retrieval of green building cases, and knowledge query. Additionally, this
study conducted a user survey using the GreenQA web platform. The results showed that 96\% of users believed the platform helped improve design efficiency. This study not only effectively supports DGBD but also provides inspiration for AI-assisted design.
\end{abstract}

\begin{figure}[h]
    \centering
    \includegraphics[width=\textwidth]{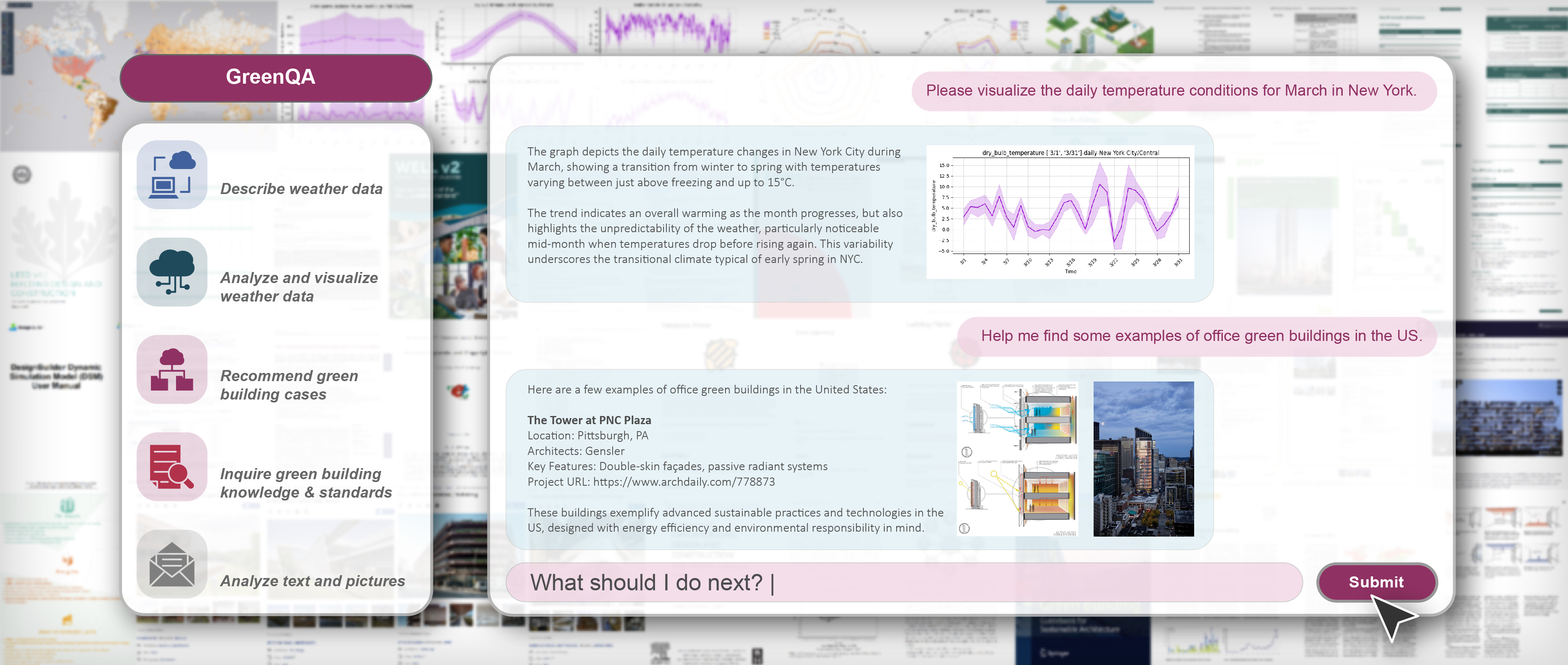}
    \caption{Conceptual illustration of the question answering platform GreenQA for decision-making in green building design.}
    \label{fig:fig1}
\end{figure}

\keywords{Design Thinking and Applied Research \and Artificial Intelligence in Design \and Green Building \and Question Answering \and Human-AI Interaction}

\section{Introduction}
The building industry accounts for up to 40\% of global greenhouse gas emissions, making a significant contribution to energy consumption \citep{DEBRAH2022104192}. During the entire lifecycle of a building, the early design stage has the potential to achieve over 40\% energy savings \citep{Han_2018}. Therefore, promoting decision-making in green building design (DGBD), which focuses on performance-oriented building design, is an important method to deal with climate change and energy shortages \citep{Doan_2017, Zhang_2017}. 

In traditional DGBD, designers primarily rely on design software for modeling and simulation software for assessment. This process depends on various complex performance parameters and extensive experience, requiring high learning costs. Additionally, the separation between simulation and optimization processes forces designers to repeat the simulation process and manually find the optimal solutions. Consequently, traditional DGBD is inefficient, prone to decision fatigue \citep{Attia_El-Degwy_Attia_2024} and lacks generalization capability in multiple scenarios \citep{Wang_Rivard_Zmeureanu_2005}. To address these issues, previous studies have leveraged artificial intelligence (AI) algorithms \citep{DEBRAH2022104192} and rapid performance analysis tools \citep{Lin_2021} to improve the efficiency of the design and simulation process. The development of large language models (LLMs) in recent years offers a new direction. The “question answering (QA)” based on natural language creates possibilities for knowledge and case retrieval \citep{Petroni_2019}, providing new ideas for the promotion and generalization of DGBD. The field of architecture is also anticipating domain-specific research based on LLMs. However, some argue that LLMs are not adept at data analysis and reasoning, with their scientific accuracy and reliability being widely questioned \citep{Lewis_2020}.

To address the issues of high learning costs and the scientific accuracy of data reasoning, as well as to further enhance design efficiency and generalization capability in the DGBD process, this study proposes an integrated question answering framework for multimodal data reasoning, named GreenQA. Based on a multimodal green building knowledge base, GreenQA leverages LLMs through new technical methods. It enables multi-turn interactive QA for various application scenarios, such as weather data analysis and visualization, case retrieval, and knowledge query. his study explores the potential applications of LLMs in the field of green building design, providing new research ideas and a technological approach to the integration of DGBD and LLMs in the future.

\section{State of the Art}

\subsection{AI in DGBD}

In recent years, artificial intelligence and machine learning have been extensively applied in the field of green building \citep{DEBRAH2022104192}, supporting green building design decision-making \citep{Petroni_2019, Li_Du_Kumaraswamy_2024}.

Case-Based Reasoning (CBR) methods are effective for retrieving knowledge from existing case libraries to improve decision-making efficiency \citep{Liu_Ma_Wang_Pan_2024}, but current databases have limitations in both quantity and quality \citep{Li_Du_Kumaraswamy_2024}, which makes it difficult to generate precise generalizations. Various methods have been explored with CBR, such as nonlinear Local-Global retrieval using Artificial Neural Networks to CBR retrieval \citep{Cheng_Ma_2015}, text mining integration \citep{Shen_2017}, and the application of CBR in building renovation \citep{Zhao_2019}, and decision-making of building envelopes design during the preliminary design stage \citep{Zhang_Li_Lin_Zhu_2021}. Additionally, Liu et al. \citep{Liu_Ma_Wang_Pan_2024} combined the Random Forest (RF) algorithm with CBR methods to address case modification issues.

Natural language processing (NLP) has also been widely applied in green building, though most studies focus on users’ perceptions and evaluations of green buildings, lacking the development of decision-making systems. Liu and Hu \citep{Liu_Hu_2019} analyzed Sina Weibo data to study public attention to green buildings, while Guo et al. \citep{Guo_2021} examined the effect of green building rating certifications on improving public satisfaction with apartments. Additionally, Du et al. \citep{Du_2021} combined NLP with green building rating systems for automatic project classification and evaluation. In the domain of building regulations, Zhong et al. \citep{Zhong_2020} integrated NLP and information retrieval to construct a building regulation QA system, but a similar integrated platform has yet to be developed in the green building field.

\subsection{QA by LLMs}
Among artificial intelligence methods, LLMs have demonstrated significant efficacy in handling complex natural language understanding and generation tasks, greatly expanding the application scope of natural language processing (NLP) \citep{Touvron_2023, Zhao_2023, OpenAI_2024}. These models, pretrained on large-scale datasets, can generate coherent and relevant text responses, and can perform NLP tasks through zeroshot or few-shot learning when given appropriate prompts \citep{Brown_2020, Chowdhery_2023}.

Question Answering (QA) is a core application area of NLP, aimed at providing accurate answers to user-posed questions. Traditional QA systems typically rely on structured knowledge bases and complex query mappings \citep{Chen_2017} to respond to user inquiries, requiring precise NLP techniques \citep{Li_2024}. With the development of LLMs, researchers have begun to explore the potential of using LLMs as knowledge bases for QA \citep{Petroni_2019}, and to compare them with traditional knowledge-based QA models \citep{Payne_2023}.

Although LLM-based QA systems can perform reasoning on open-ended questions, general LLMs often lack specialized knowledge in the field of green building. This can lead to the generation of answers that are coherent in natural language but factually incorrect \citep{Kim_2023, Zhuang_2023}, known as hallucination \citep{Ji_2023}. This limitation significantly constrains the support that LLMs can provide in the field of green building.

\section{Methods}
\begin{figure}[ht]
    \centering
    \includegraphics[width=\textwidth]{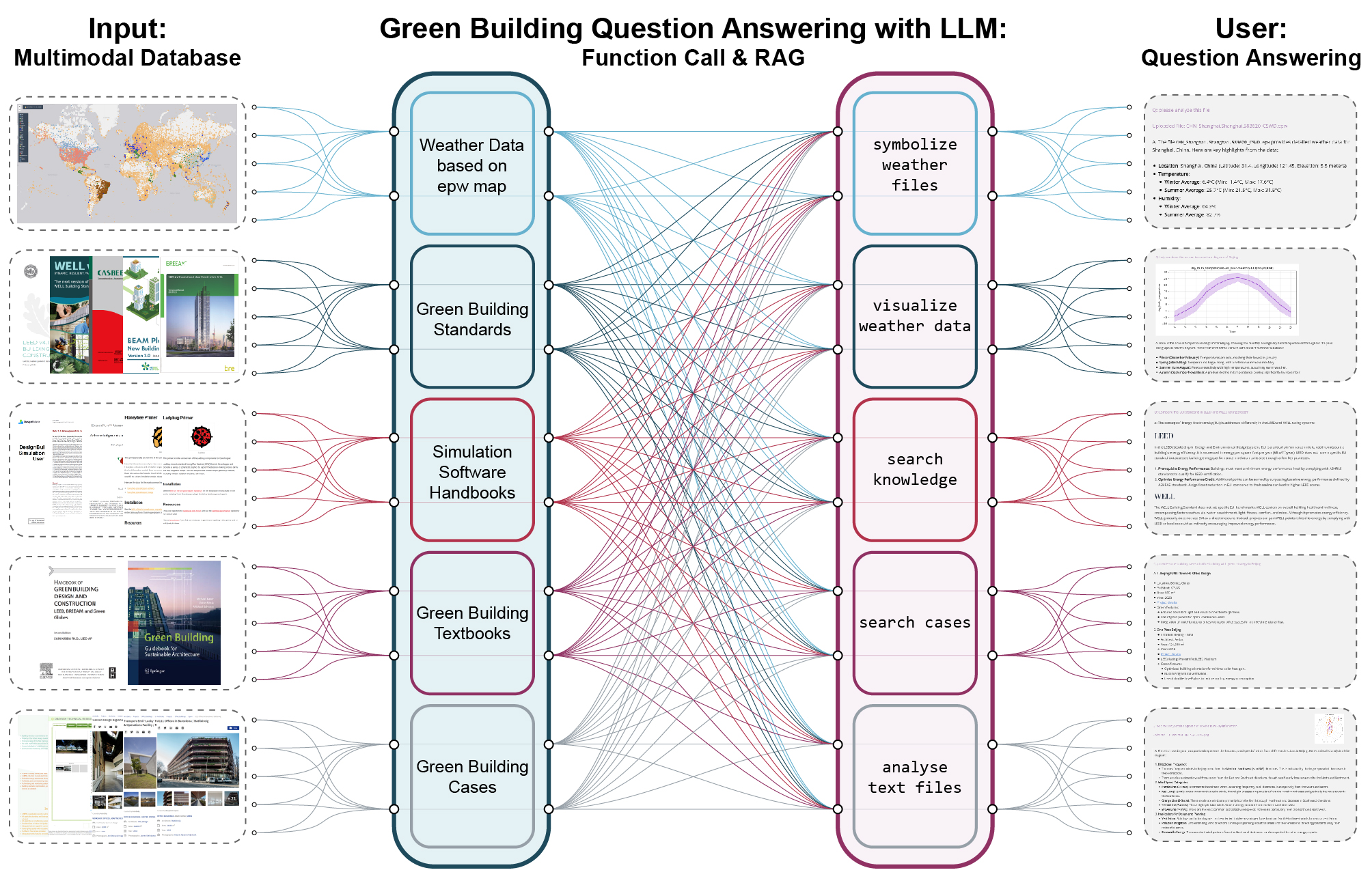}
    \caption{Research framework.}
    \label{fig:fig2}
\end{figure}

This study developed the GreenQA platform, which supports knowledge retrieval and design interaction in green building design. The technical framework of the platform is illustrated in Figure \ref{fig:fig2}, and its construction is based on a multimodal green building knowledge base. 

To address various QA needs in DGBD scenarios, this study enhances the performance of LLMs in three aspects and restructures the output response mechanism. First, by using Retrieval Augmented Generation (RAG) to automatically connect the knowledge base and prevent hallucinations. Second, by employing Chain of Thought (CoT) to enhance LLMs’ contextual memory and reasoning capabilities. Third, by utilizing Function Call to integrate LLMs with external interfaces, based on the previous two technologies, to meet the demands of more complex data reasoning tasks and support the reading and generation of documents, images, tables, and other modalities.

Integrating these technical methods, an online QA platform, GreenQA, was developed on the web. A user survey was conducted to evaluate its performance and identify areas for improvement.

\subsection{Knowledge Base Preparation}

DGBD relies on multimodal background knowledge, encompassing technical indicators like acoustics, lighting, and thermal performance, which have led to the creation of numerous textbooks and standards. Globally, up to 600 green building rating systems have been established to assess sustainability \citep{Doan_2017, Zhang_2017, Awadh_2017}. Notably, the UK's BREEAM and the US's LEED standards have been pivotal in advancing sustainable buildings. \citep{Lee_2013}. Research indicates that BREEAM and LEED-certified buildings consume approximately 30\% less energy than non-certified buildings \citep{Doan_2017}. The application of these evaluation standards has accumulated numerous cases and practical experiences, which can be leveraged to address DGBD challenges. 

\begin{figure}[ht]
    \centering
    \includegraphics[width=\textwidth]{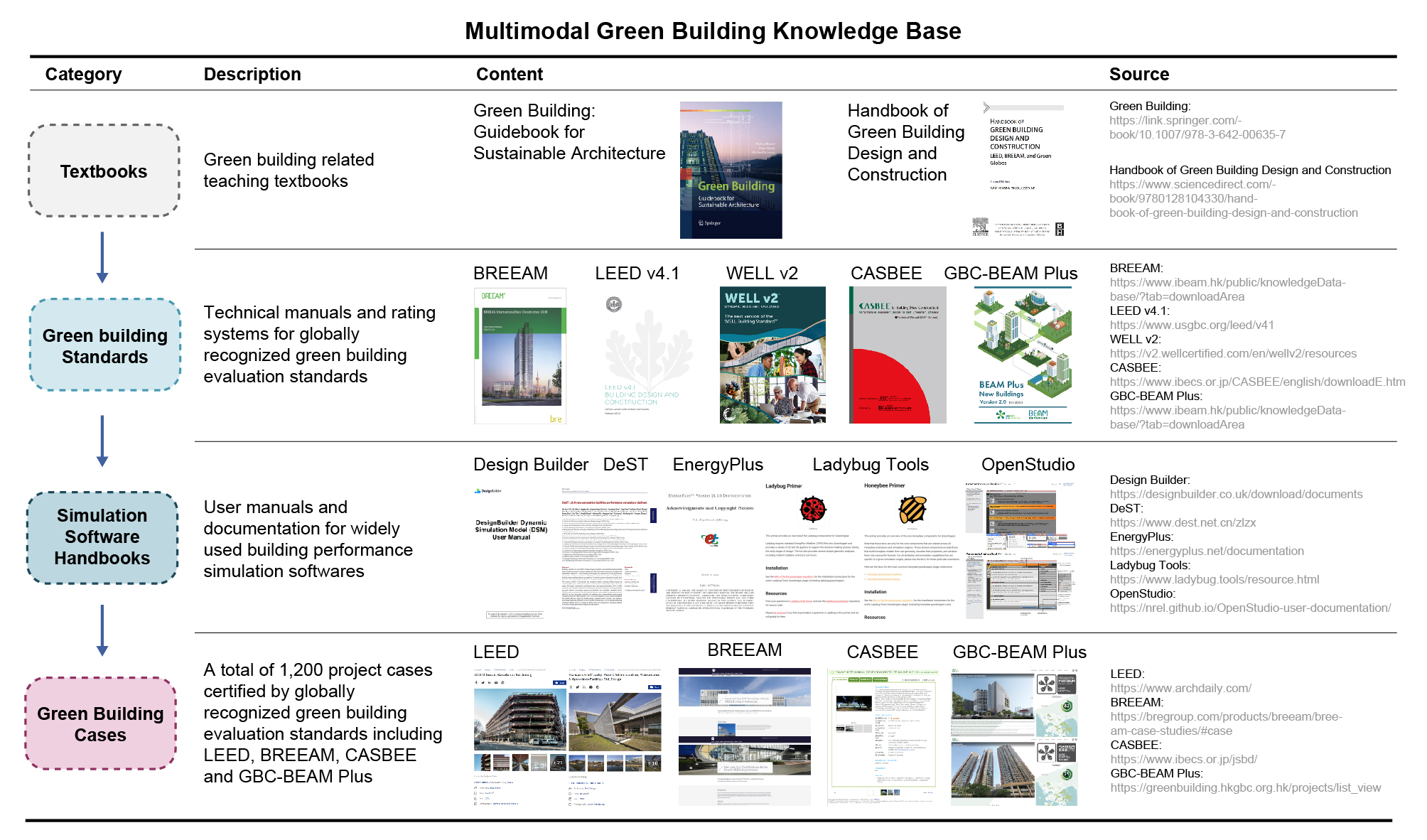}
    \caption{Multimodal green building knowledge base.}
    \label{fig:fig3}
\end{figure}

Based on this logic, we constructed a multimodal green building knowledge base comprising four components: green building theoretical textbooks, green building rating standard guidelines, performance simulation software manuals, and green building cases. The specific contents of the knowledge base are illustrated in Figure \ref{fig:fig3}. The books, standards, and manuals are sourced from their respective official websites.

The green building cases include certified projects from four green building rating systems, taking the collection and cleaning of LEED cases as an example. First, the process begins with using web scraping tools to gather basic information and text descriptions of LEED-certified projects from the ArchDaily website. Next, the OpenAI API is called to process the natural language in the case texts, extracting sentences related to LEED ratings and building performance. The project information and relevant descriptions are then organized into structured JSON files.

\subsection{QA by RAG}
Due to the lack of domain-specific knowledge bases in general LLMs, they tend to generate hallucinations or incorrect answers when handling knowledge-intensive tasks. Relevant research has proposed the Retrieval Augmented Generation (RAG) method to improve the accuracy of LLMs in QA tasks \citep{Lewis_2020}. In knowledge-intensive industries such as law and medicine, studies have already demonstrated the use of the RAG method to provide more accurate answers with the aid of knowledge bases \citep{Louis_Dijck_Spanakis_2024, singhal2023}. However, in the field of architecture, a systematic database and methodology have not yet benn developed.

To further achieve the customized application of LLMs in assisting DGBD, this study adopts the RAG method. By calculating semantic similarity, relevant information is retrieved from the external knowledge base to enhance the expertise of LLMs in the field of green building, while also ensuring the timeliness of the knowledge base \citep{Chen_Lin_Han_Sun_2024, Guu_2020, Lewis_2020}.

\begin{figure}[ht]
    \centering
    \includegraphics[width=\textwidth]{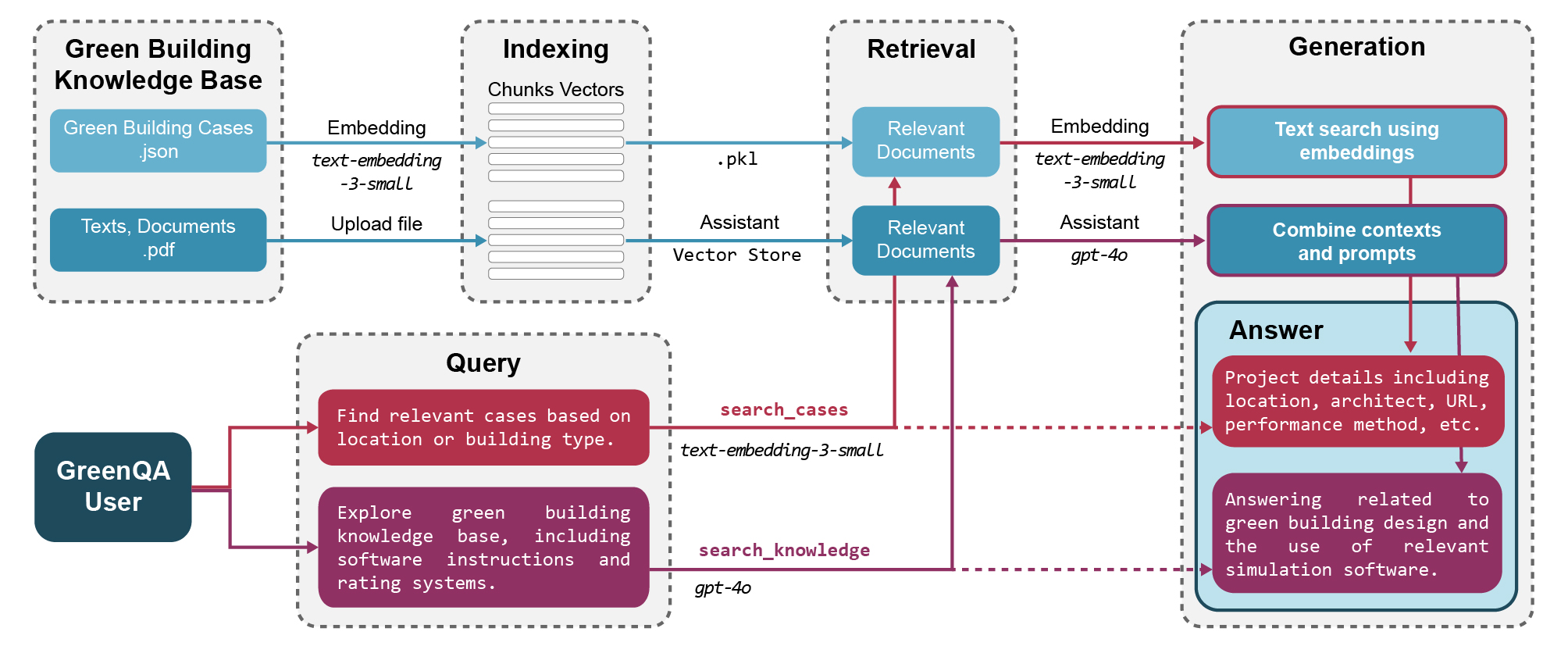}
    \caption{Implementation of cases and knowledge base retrieval.}
    \label{fig:fig4}
\end{figure}

The RAG method in GreenQA is based on the basic framework of indexing, retrieval, and generation \citep{Gao_2024}, as illustrated in Figure \ref{fig:fig4}. For each user query, the system reads text data stored in the knowledge base as vectors, calculates the vector similarity between the user’s question and the indexed resources, and then generates a response based on LLM. Specifically, green building cases are stored locally as vector files through an embedding model, and calculations and retrievals are then performed for each user query. This method significantly reduces the number of tokens required for each QA session while markedly improving response speed.

\subsection{QA by CoT}

QA tasks for DGBD often involve data reasoning processes with complex interaction logic. Single-turn QA typically results in low answer accuracy for these challenging tasks (e.g., arithmetic and commonsense reasoning). Increasing the model’s parameter scale does not significantly enhance performance for such tasks \citep{Lewis_2020}. Therefore, we must select a more appropriate technical approach.

\begin{figure}[ht]
    \centering
    \includegraphics[width=\textwidth]{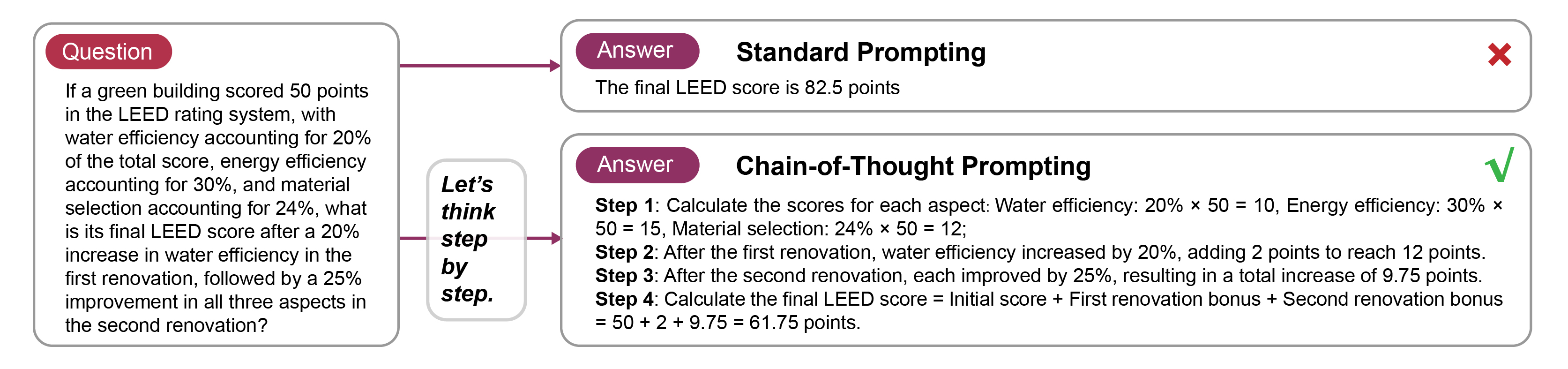}
    \caption{Comparison of results with/without using zero-shot CoT.}
    \label{fig:fig5}
\end{figure}

In 2022, Wei et al. proposed using Chain of Thought (CoT) to solve this problem \citep{Wei_2022}. CoT guides LLMs to break down complex tasks by constructing example prompts, significantly improving task accuracy. This study employs the zero-shot CoT method, incorporating specific instructions like “Let’s think step by step” into the prompts (Figure \ref{fig:fig5}). This method is straightforward and achieves significant performance improvements for LLMs with more than 100 billion parameters \citep{Kojima_Gu_Reid_Matsuo_Iwasawa_2022}.

\begin{figure}[ht]
    \centering
    \includegraphics[width=0.5\textwidth]{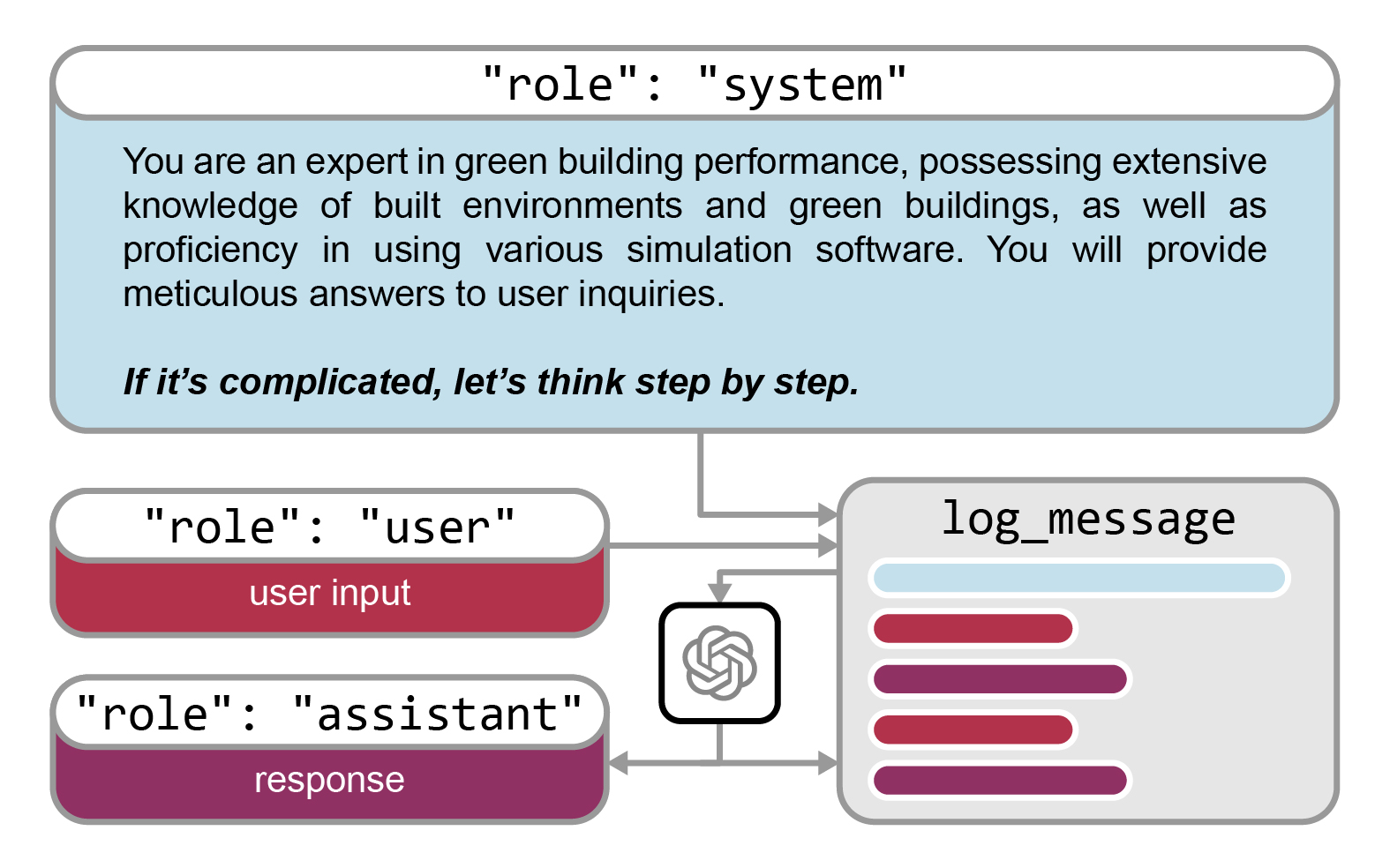}
    \caption{Implementation of multi-turn dialogue based on CoT.}
    \label{fig:fig6}
\end{figure}

To achieve a more continuous interactive experience, we use a loop structure to construct multi-turn conversations, allowing the model to answer subsequent questions based on previous context, which effectively helps users complete complex tasks. This study uses the latest GPT-4o API provided by OpenAI as the QA model \citep{OpenAI_2024_update}. Figure \ref{fig:fig6} demonstrates the multi-turn QA framework constructed with this model using CoT.

\subsection{QA by Function Call}

In real-world DGBD scenarios, EPW (EnergyPlus Weather File) files are crucial for understanding the relationship between buildings and the environment through performance simulation analysis \citep{Crawley_Hand_Kummert_Griffith_2008}. However, directly inputting EPW files and user requests (such as “\textit{Please help me analyze this EPW weather data file.}” or \textit{“What type of buildings are suitable for the temperature and humidity in this area?}”) into LLMs would consume a vast number of tokens, and the probabilistic generation of direct inference results significantly reduces the credibility of data reasoning.

\begin{figure}[ht]
    \centering
    \includegraphics[width=\textwidth]{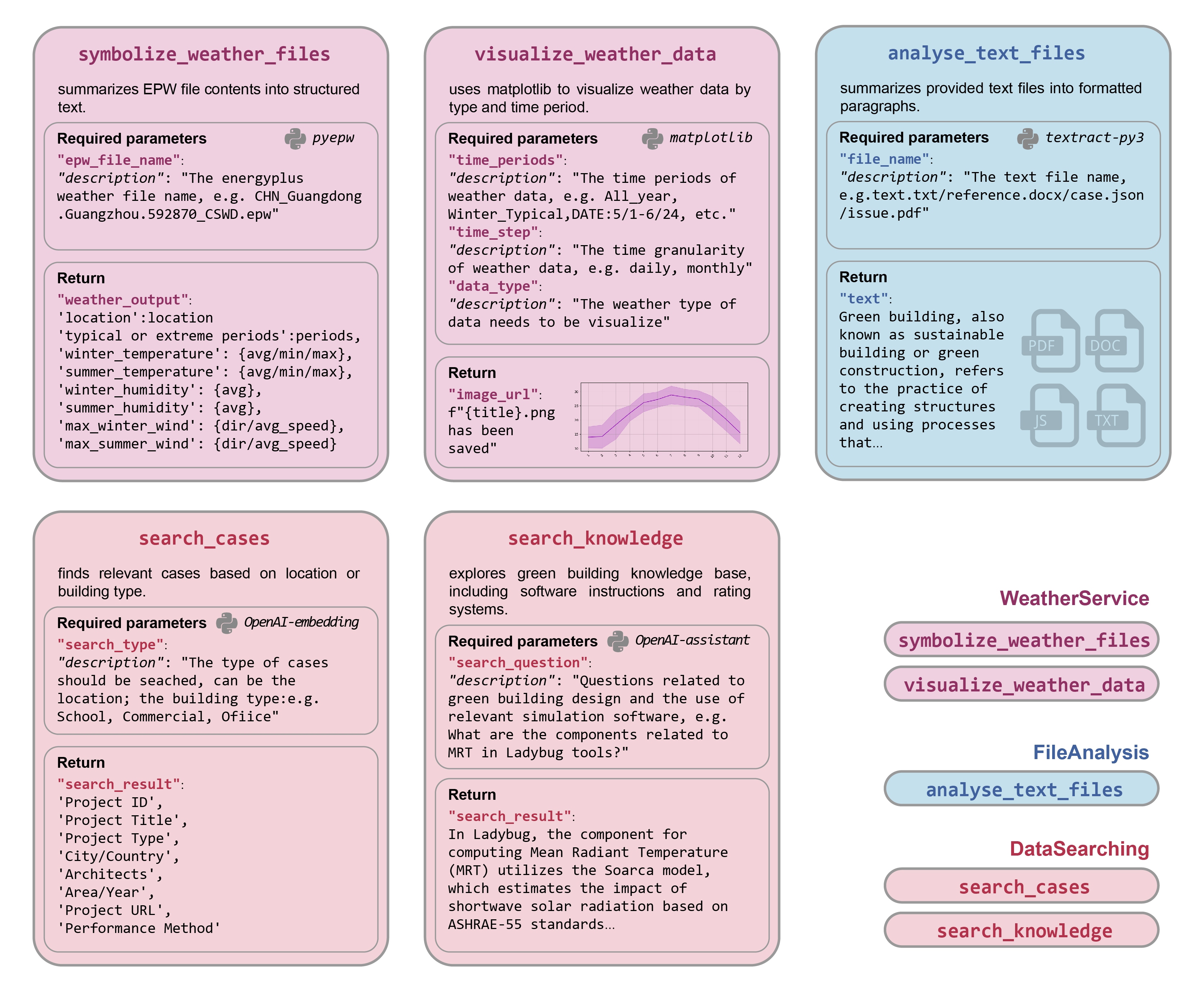}
    \caption{Elaboration of functions and corresponding input/output.}
    \label{fig:fig7}
\end{figure}

This study uses the Function Call feature to address the above problem. Function Call originated from agent tools in LLM research. In the commonly used LLM application framework LangChain \citep{Harrison_2022}, users can create agent scripts based on their needs to extend the LLM’s capabilities, particularly its data reasoning functions \citep{Topsakal_Akinci_2023}. OpenAI’s recent releases have integrated this feature into their API \citep{OpenAI_2023}, allowing users to create a function with corresponding input and output descriptions in natural language and a real function script on the back-end. When the LLM processes the user’s question, it determines whether to call the function based on the function’s description. We have constructed five functions to achieve rich multimodal complex interactions. Figure \ref{fig:fig7} shows the functions’ capabilities and input-output content.

\begin{figure}[ht]
    \centering
    \includegraphics[width=\textwidth]{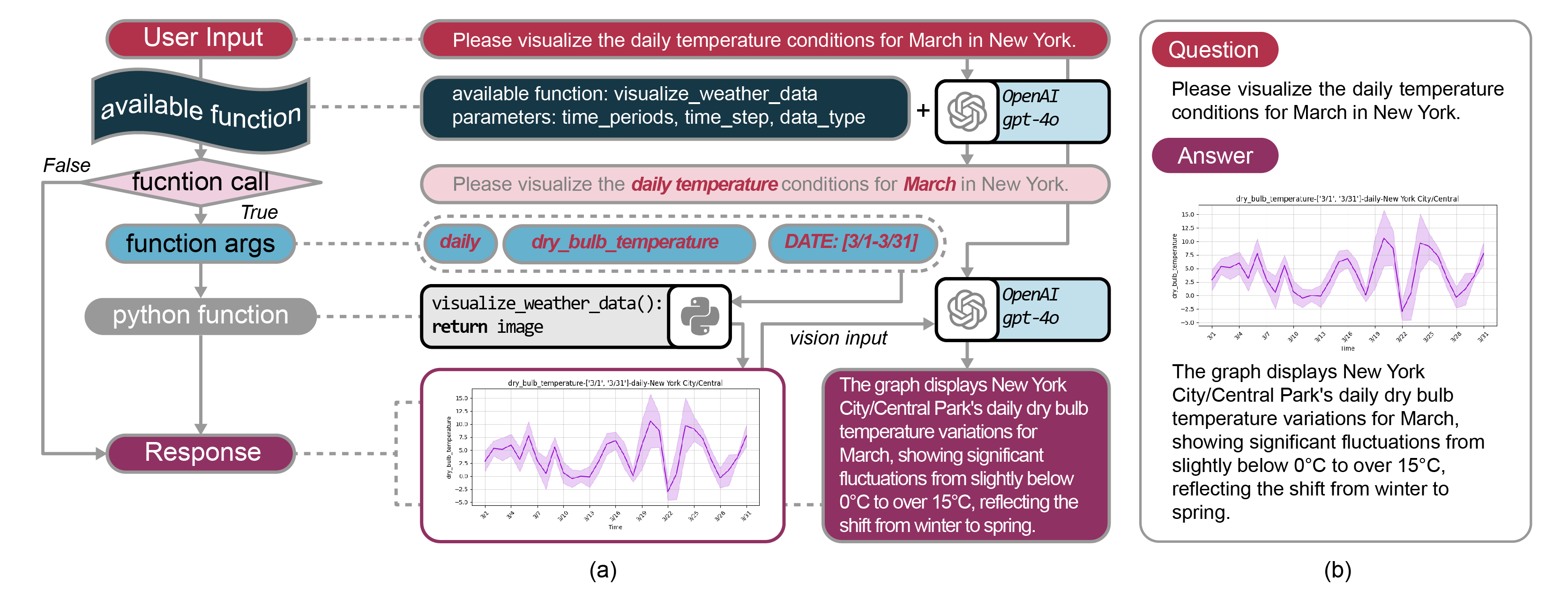}
    \caption{(a) Function call implementation; (b) Function call QA example.}
    \label{fig:fig8}
\end{figure}

Figure \ref{fig:fig8} exemplifies the Function Call process using the visualize\_weather\_data function, demonstrating a potential QA scenario in green building design to better explain the Function Call workflow.

\begin{enumerate}[font=\footnotesize] 
\item Describe the function and its required input parameters in natural language in the tools, extracting the EPW file information as formatted data and using the Matplotlib library to plot meteorological data charts based on user input needs, thereby avoiding direct LLM handling of complex data.
\item Embed a Python function in the back-end that matches the description in the tools for actual data processing, and set it as an executable function in the API’s available\_functions.
\item Read user input to determine whether to call the predefined function, e.g., “Please visualize the daily temperature conditions for March in New York.”
\item Interpret the user intent, automatically extract and convert the text to the input format specified for the designed function, such as time\_step="daily", time\_periods="DATE:3/1-3/31", data\_type="dry\_bulb\_temperature", and pass these parameters to the corresponding Python script.
\item Return the Function Call output, including text and images, as the response, and combine it with the initial query to form a new turn of user input to generate the final response.
\end{enumerate}

\subsection{Platform Construction}

Based on the web platform, this study has developed an interactive website called GreenQA, which uses the technical framework shown in Figure 9 to handle user input and output information flow, enabling knowledge inference QA related to the field of green building.

\begin{figure}[ht]
    \centering
    \includegraphics[width=\textwidth]{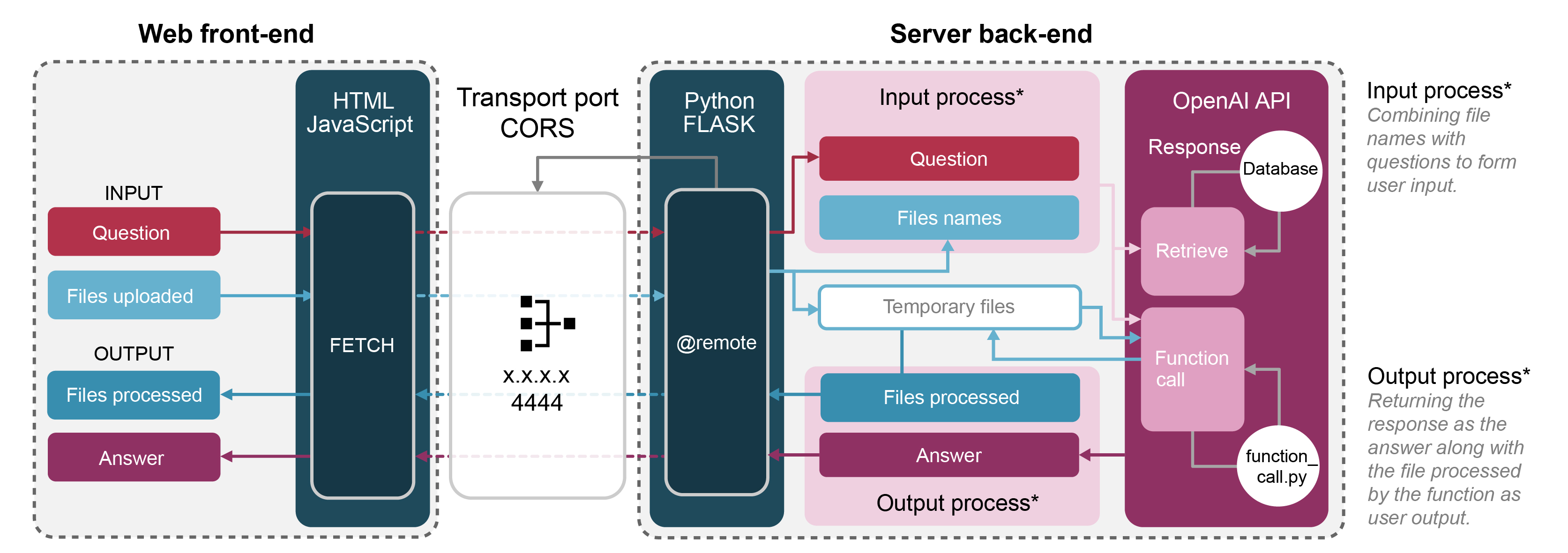}
    \caption{Interaction logic of GreenQA platform front-end \& back-end.}
    \label{fig:fig9}
\end{figure}

The back-end of the website utilizes the Fetch mode in JavaScript to send responses, which are received by a Python Flask framework and executed by corresponding Python scripts. Additionally, enabling CORS mode allows these scripts to process cross-origin requests.

The dialog API processing section constructs a complex yet efficient information transmission logic. When the input information flow reaches the back-end, user-uploaded files are saved in a local temporary folder, and the file names are appended to the user input text before being sent to the API as input. This approach allows the Function Call feature to recognize the corresponding file names and call the required functions. When the API generates responses and function calls create charts or other files, these files are separated from the text, saved in the temporary folder, and then sent to the front-end. This method facilitates multimodal interaction and simplifies the recording of information transmission results.

\begin{figure}[ht]
    \centering
    \includegraphics[width=0.9\textwidth]{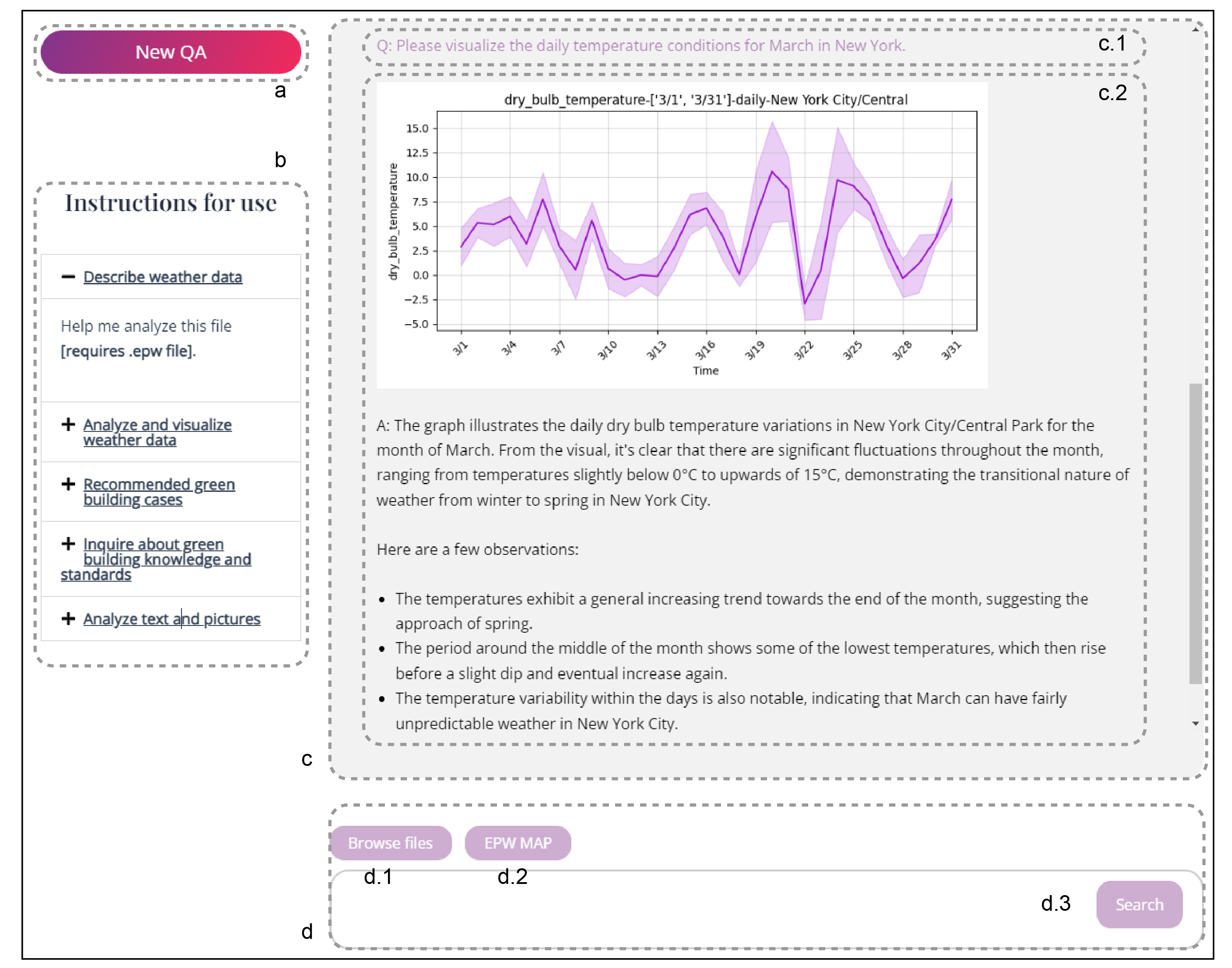}
    \caption{Website page layout.}
    \label{fig:fig10}
\end{figure}

Referring to common QA websites such as ChatGPT, the platform is deployed on a front-end managed by WordPress using HTML and JavaScript markup languages as shown in Figure \ref{fig:fig10}. Users can enter their questions in the search box d.3 located at the bottom of the page. Above the search box, there are two buttons: Button d.1 allows users to upload multiple files for the LLM to read, supporting multimodal file formats such as EPW files, images (.jpeg/.jpg/.png), and text files (.txt/.json/.pdf/.docx). The second button d.2 is used to open the EPW Map global meteorological data visualization website built by Ladybug Tools \citep{Roudsari_Nunzio_Peng_2023}, which helps users find and download EPW data for specific regions. After clicking the search button located at the bottom right, a dialog response will be generated, and the response results will be displayed in the board c above the search box. Clicking button a at the top left of the page, labeled "New QA," will start a new round of dialogue. The lower left corner b provides some QA examples.

\subsection{User Study}

To better understand user satisfaction and gather suggestions for platform improvement, a survey was conducted, as detailed in the appendix. The questionnaire was designed to evaluate four aspects: multi-turn QA and information interaction, multimodal data and information inference, the value and practicality of key functions, and changes in work efficiency.

The participants in this survey were senior undergraduate and graduate students with previous experience in building performance simulation software or green building design. Before completing the questionnaire, each participant was required to use the platform in a real design scenario for 30 to 60 minutes. Scoring was based on a 5-point scale.

\section{Result and Discussion}

\subsection{Results of Knowledge Base Construction}

\begin{figure}[ht]
    \centering
    \includegraphics[width=0.5\textwidth]{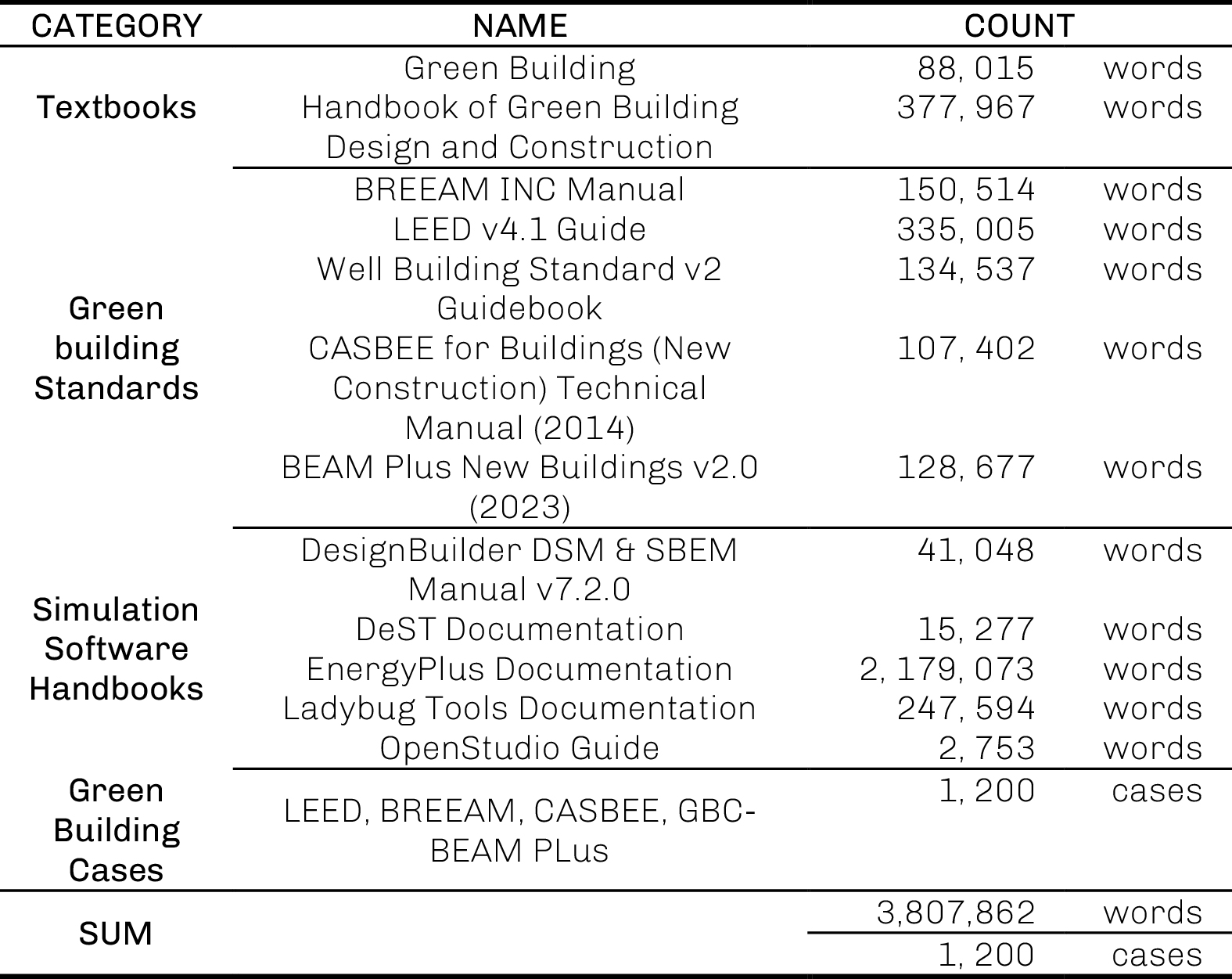}
    \caption{Statistical results of green building knowledge base.}
    \label{fig:fig11}
\end{figure}

\begin{figure}[ht]
    \centering
    \includegraphics[width=\textwidth]{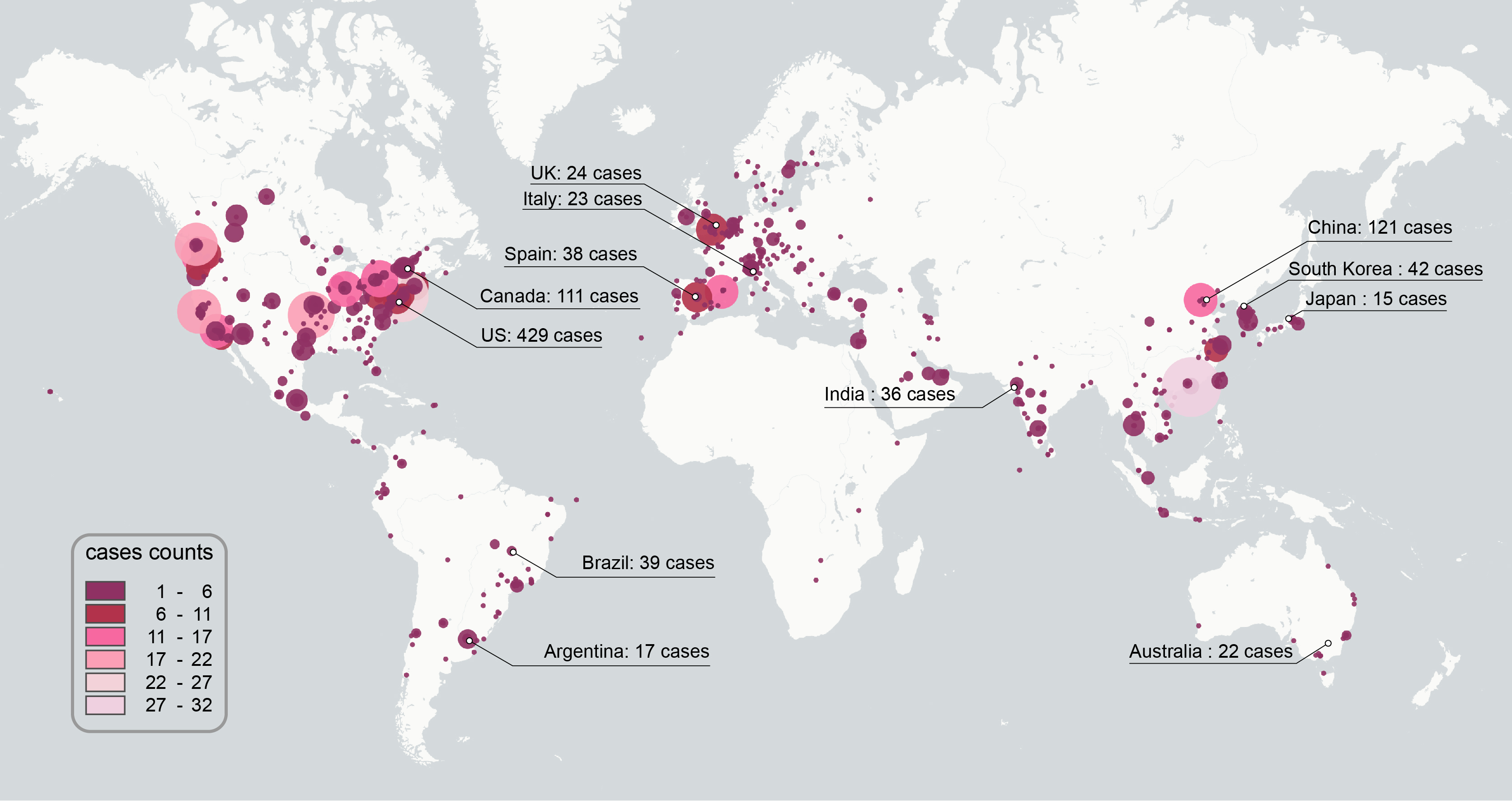}
    \caption{Global distribution of green building cases.}
    \label{fig:fig12}
\end{figure}

The multimodal green building knowledge base has amassed over 3.8 million words of textual materials and includes 1,200 green building cases (Figure \ref{fig:fig11}). These cases span multiple countries and regions across all continents. As shown in Figure \ref{fig:fig12}, the highlighted countries are those with the highest number of cases in each continent.

\begin{figure}[ht]
    \centering
    \includegraphics[width=0.5\textwidth]{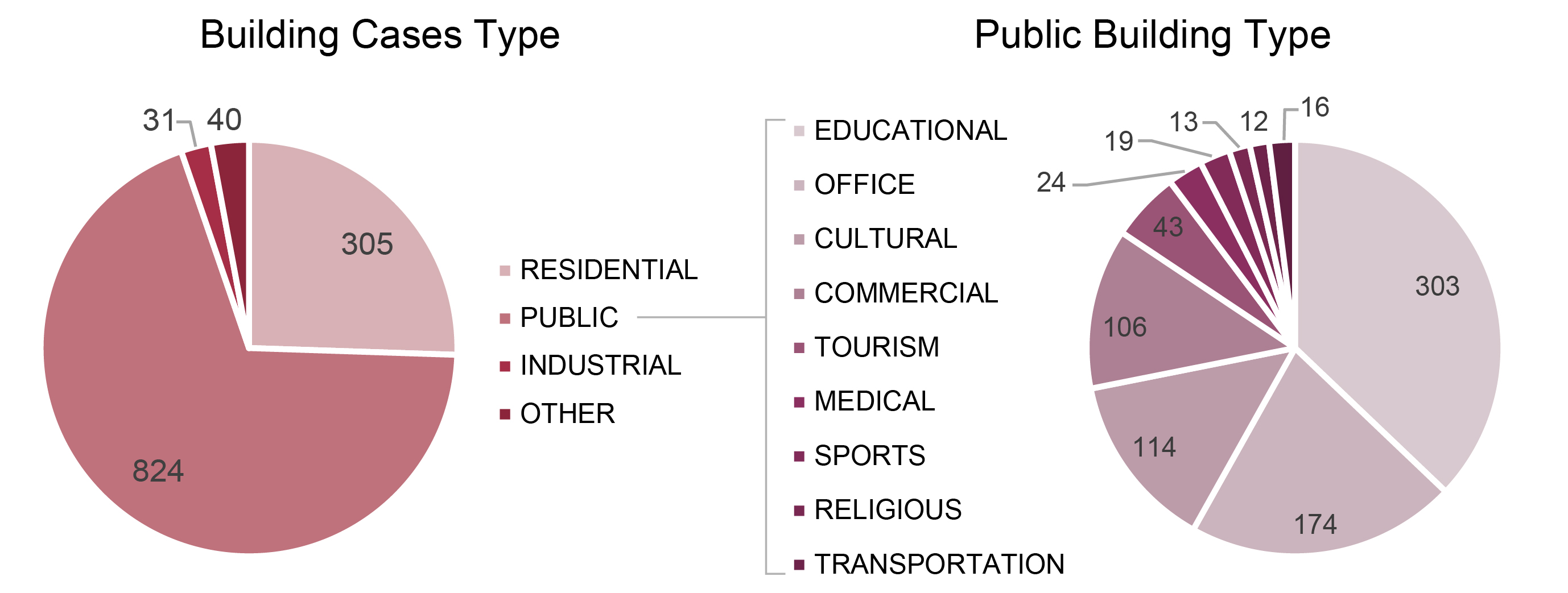}
    \caption{Type and quantity of green building cases.}
    \label{fig:fig13}
\end{figure}

The type and quantity of projects for green building cases are shown in Figure \ref{fig:fig13}. The main categories include residential, public, and industrial types. Within the public building category, there are subtypes such as educational, office, and cultural buildings.

\subsection{Results of Interactive QA}

This study uses a simulated interactive case to evaluate the
platform’s QA results. 

In a simulated design scenario, users engage with the platform to address fundamental questions related to green building concepts, acquire local weather data in EPW format for performance simulations, and integrate data into preliminary architectural designs. The platform assists in extracting key information and generating visualizations, supporting users throughout the modeling and simulation
process to address specific design queries and local regulations.

Figure \ref{fig:fig14} illustrates specific QA content on the platform for various scenarios encountered in the simulated process. It can be observed that, for 1) single-task questions, 2) questions requiring Multi-Function Call, and 3) questions requiring Function Multi-Call with different parameters, this platform consistently delivers professional and rigorous outputs, assisting users in engaging in practical design tasks.

\subsection{Results of Survey}

\begin{figure}[ht]
    \centering
    \includegraphics[width=\textwidth]{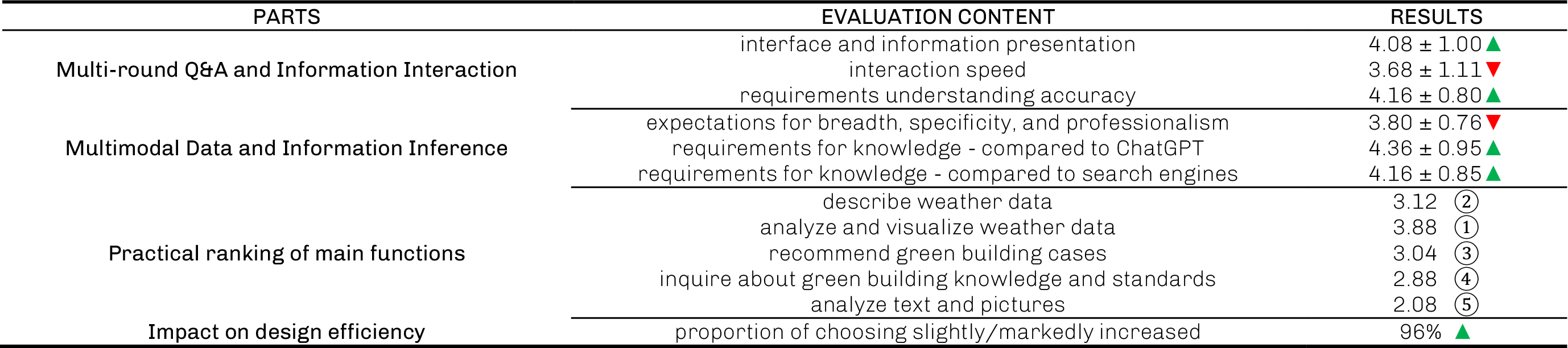}
    \caption{Statistical results of user survey.}
    \label{fig:fig15}
\end{figure}

Figure \ref{fig:fig15} presents the statistical results of the survey. Among the 25 collected questionnaires, users expressed an overall high level of satisfaction with the GreenQA decision support platform. They particularly appreciated the clear and precise presentation of information as well as the accurate understanding of their needs. Most users found the platform highly valuable for both analyzing meteorological data and retrieving building case studies. Additionally, 96.0\% of users reported an increase in work efficiency (scoring above 4) after using the GreenQA platform to assist with DGBD.

However, users gave relatively lower ratings (below 4) for the platform’s interaction speed as well as for its robustness and professionalism.

\subsection{Discussion}

\begin{figure}[ht]
    \centering
    \includegraphics[width=\textwidth]{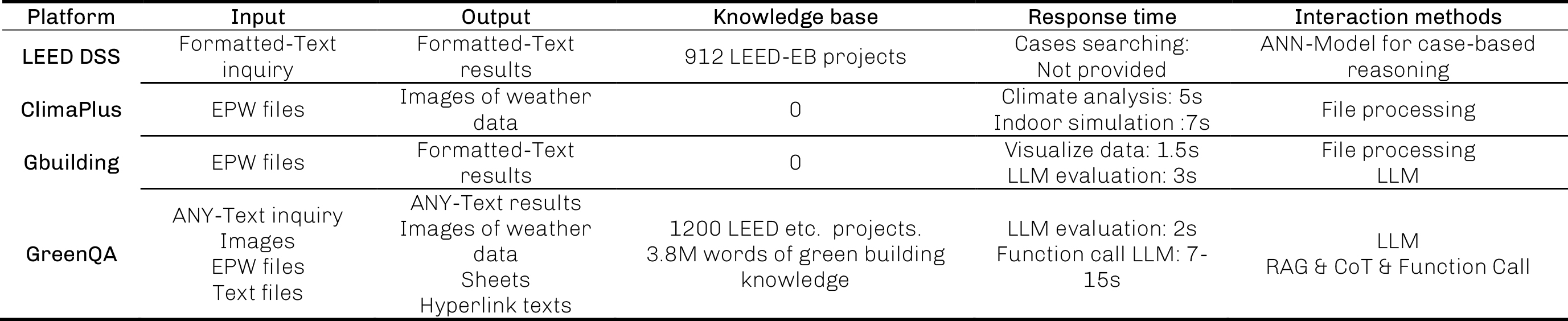}
    \caption{Comparison with other platforms.}
    \label{fig:fig16}
\end{figure}

\begin{figure}[h]
    \centering
    \includegraphics[width=0.8\textwidth]{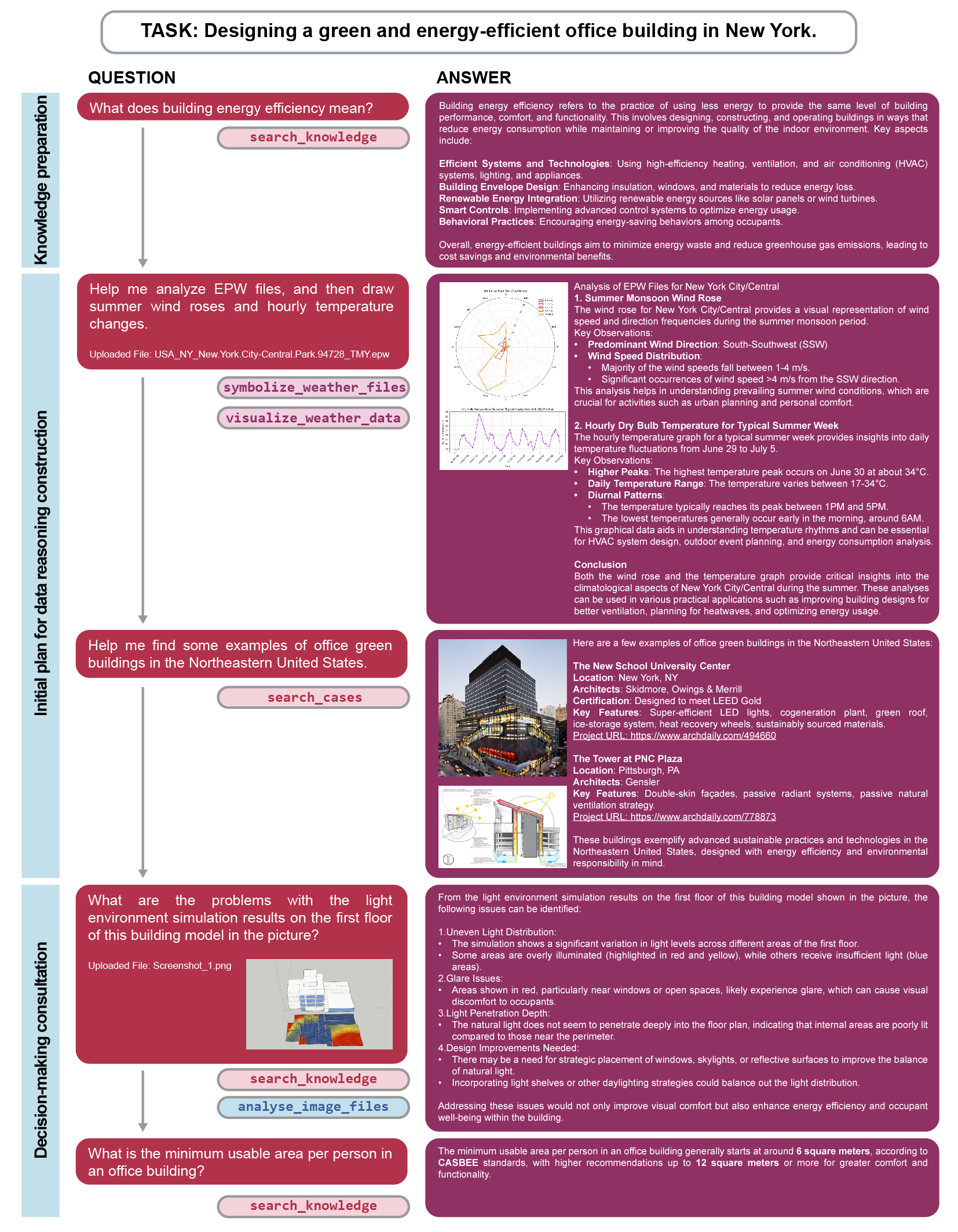}
    \caption{A simulated using example of GreenQA platform.}
    \label{fig:fig14}
\end{figure}

This study also compared this platform with other platforms supporting DGBD and weather analysis, such as LEED DSS \citep{Jun_Cheng_2017}, ClimaPlus \citep{Arsano_Reinhart_2020}, and Gbuilding \citep{Zhen}. We examined the input/output content of these platforms and evaluated them based on dimensions such as knowledge base data volume, single input/output response time, and response output mechanisms, as shown in Figure \ref{fig:fig16}.

The comparison shows that the GreenQA platform exhibits significant advantages in terms of functionality, knowledge base size, response time, and interaction methods. GreenQA supports various input and output types, including arbitrary text queries, images, EPW files, and text files, enabling it to handle diverse data types. Additionally, GreenQA possesses an extensive knowledge base containing 1,200 projects and 3.8 million words of green building knowledge, providing comprehensive and accurate information outputs. In terms of response time, it balances flexibility and speed. Furthermore, GreenQA is built on LLM and is compatible with RAG, CoT, and Function Call, offering an innovative and efficient design interaction experience.

Despite positive survey results, some shortcomings remain in the platform’s knowledge generalization, interaction performance, and accuracy. Firstly, the green building knowledge base lacks comprehensive coverage, with gaps in green building standards for certain regions, limited image and video materials, and a lack of representative cases from extreme climate zones, such as parts of Africa and the polar regions. Secondly, the platform has relatively low operational redundancy, with API-based language model calls being especially time-consuming when searching the knowledge base. Lastly, although the framework avoids most hallucinations through retrieval augmentation and case-based reasoning, the generative model may still fabricate parameters for Function Call when dealing with more complex and ambiguous questions.

\section{Conclusion}

As a new paradigm of AI, LLMs are reshaping our work patterns and understanding of the world. Their interaction capabilities are poised to transform design processes, with “Design with LLMs” likely becoming the mainstream approach in computer-aided design in the future. However, significant issues remain, including information accuracy, interpretability, and data security.

This study not only effectively supports DGBD but also redefines traditional design paradigms. Firstly, it establishes a multimodal database containing 3.8 million words of green building knowledge and 1,200 green building cases, filling the gap in large-scale multimodal databases in the field of green building. Secondly, leveraging LLMs' interactive abilities, the study interprets natural language instructions and provides precise responses to DGBD questions. Lastly, by integrating RAG, CoT and Function Call, it enables precise knowledge retrieval and data-model separation, reducing hallucinations and data leakage, and enhancing the professionalism of LLM-based QA in the field of green building.

This study also highlights future directions for the method: (1) localizing large models and expanding the knowledge base, (2) optimizing interaction processes and enhancing search response speed, 3) increasing fault tolerance for user requirements. This will support more complex and ambiguous design decisions, contributing to energy saving and sustainability in green building design.

\section*{Acknowledgments}
To improve the clarity and grammatical accuracy of the textural content, we utilized the capabilities of ChatGPT [https://chat.openai.com/] for copy-editing purposes only.

This study is supported by National Key R\&D Program of China (2022YFC3801301), the National Natural Science Foundation of China (Grant No. 52130803, 52394223).

Thank you to Zixu Zhen for providing study models and valuable inspiration during the early stages of the research.


\bibliographystyle{apalike}
\bibliography{main} 

\typeout{get arXiv to do 4 passes: Label(s) may have changed. Rerun}

\clearpage
\appendix
\section{Appendix}
\subsection*{User Survey Questionnaire}

\begin{enumerate}
    \item Multi-turn QA and Information Interaction
    \begin{itemize}
        \item Do you find the webpage interface and information presentation of GreenQA clear, concise, and easy to use?
        \item Are you satisfied with the interaction speed of GreenQA?
        \item Do you think GreenQA can accurately understand your question requirements?
    \end{itemize}

    \item Multimodal Data and Information Inference
    \begin{itemize}
        \item Do you think the answers provided by GreenQA meet your expectations for breadth, specificity, and professionalism?
        \item Compared to directly using ChatGPT, do you think the GreenQA platform can more accurately meet your needs for knowledge about green buildings?
        \item Compared to manually searching for relevant data, do you think the GreenQA platform can more integrately and conveniently meet your needs for knowledge about green buildings?
    \end{itemize}

    \item How would you rank the value and practicality of the main functions of GreenQA?
    \begin{itemize}
        \item Describe weather data
        \item Analyze and visualize weather data
        \item Recommend green building cases
        \item Inquire about green building knowledge and standards
        \item Analyze text and pictures
    \end{itemize}

    \item How do you think using the GreenQA platform to assist in green building design decision-making has changed your work efficiency?
    \begin{itemize}
        \item Markedly decreased.
        \item Slightly decreased.
        \item Remained the same.
        \item Slightly increased.
        \item Markedly increased.
    \end{itemize}
\end{enumerate}

\end{document}